\documentclass{llncs}
\usepackage{multirow}
\usepackage{graphicx}
\graphicspath{ {./images/} }
\usepackage{float}
\usepackage{amsmath,amssymb} % define this before the line numbering.
\usepackage{color}
\usepackage[width=135mm,left=23mm,paperwidth=176mm,height=210mm,top=18mm,paperheight=246mm]{geometry}
\usepackage{hyperref} 
\usepackage{cleveref}
\usepackage{caption,subcaption}
\usepackage[bottom]{footmisc}

\begin{document}
% \renewcommand\thelinenumber{\color[rgb]{0.2,0.5,0.8}\normalfont\sffamily\scriptsize\arabic{linenumber}\color[rgb]{0,0,0}}
% \renewcommand\makeLineNumber {\hss\thelinenumber\ \hspace{6mm} \rlap{\hskip\textwidth\ \hspace{6.5mm}\thelinenumber}}
% \linenumbers
\pagestyle{headings}

\mainmatter

\title{Detection of Driver Drowsiness by Calculating the Speed of Eye Blinking} % Replace with your title

%\titlerunning{URAI2021}
%\authorrunning{Max Mustermann}

\author{Muhammad Fawwaz Yusri \and Patrick Mangat \and Oliver Wasenmüller}
\institute{Mannheim University of Applied Science, Germany\\ 
\email{muhammadfawwaz.yusri@stud.hs-mannheim.de} \\
\email{p.mangat@hs-mannheim.de} \\
\email{o.wasenmueller@hs-mannheim.de}
}

\maketitle

\begin{abstract}
Many road accidents are caused by drowsiness of the driver. While there are methods to detect closed eyes, it is a non-trivial task to detect the gradual process of a driver becoming drowsy. We consider a simple real-time detection system for drowsiness merely based on the eye blinking rate derived from the eye aspect ratio. For the eye detection we use HOG and a linear SVM. If the speed of the eye blinking drops below some empirically determined threshold, the system triggers an alarm, hence preventing the driver from falling into microsleep. In this paper, we extensively evaluate the minimal requirements for the proposed system. We find that this system works well if the face is directed to the camera, but it becomes less reliable once the head is tilted significantly. The results of our evaluations provide the foundation for further developments of our drowsiness detection system.  

\end{abstract}

\section{Introduction}
Around 74\% of European road users mostly agree that tired driving or microsleep is a frequent crash cause. The statistics were gained in 2018 by E-Survey of road users’ attitudes from more than 35,000 respondents across 32 countries \cite{Charles}. 

Thus, driver monitoring becomes of increased importance \cite{sviro}, since the consequence of drowsiness can be recognized distinctively during driving. This behavior can be seen as the driver slowly starts losing consciousness. Furthermore, one of the important characteristics of drowsiness is slow eye movement \cite{parisa,Tiago}. In this paper, the movement of the eyes will be the key criterion to distinguish between wakeful and drowsy drivers. 
We implement and evaluate a practical and simple drowsiness detection algorithm that can be easily integrated into driver-assistance systems. The system is merely based on the eye aspect ratio and eye blinking rate, where we combine Histograms of Oriented Gradients (HOG) and linear support vector machines for reliable and accurate eye detection. Upon extensive experiments, we determine a threshold for the eye blinking rate, below which our algorithm triggers an alarm. We conducted extensive evaluations based various test cases, which challenge our system. While our drowsiness detection algorithm works in principle, we identify circumstances, in which our system is less accurate. In this way we systematically elaborate the next steps to further improve our simple drowsiness detection system.
%To achieve better results in calculating the speed of eye blinking, we have conducted extensive evaluations while taking different scenarios into account. In this way we can systematically elaborate the next steps to further improve our simple drowsiness detection system.

%TODO: In the last paragraph we should rather briefly summarize the main goal of our paper. The story could be: we want to install a simple drowsiness detection merely based on EAR and eye blinking rate, and provide extensive evaluation.  
%Apart from other approaches in detecting driver’s drowsiness, we have realized a practical and simple solution that can be integrated into the car’s system. The system is merely based on EAR and eye blinking rate. To achieve better results in calculating the speed of eye blinking, we have conducted extensive evaluations while taking different scenarios into account. In this way we can systematically elaborate the next steps to further improve our simple drowsiness detection system.
%This paper is divided into five sections. The first section covers the introductions and followed by related works in section two. The third section describes the methodology of this paper. Section three is the evaluations from different tests and the last section is the conclusion.}

\section{Related Work}

There are several methods to detect the features of the eyes as well as drowsiness. For instance, some of the researches apply Viola-Jones cascade classifier to differentiate the eyes from other facial parts \cite{Zeeshan,Arafat}.  By determining the number of pixels on the iris, cornea and eyelid, the number of blinking and duration of the closed eyes can be calculated. While comparing the number of blinking with the blink rate set (normal $=$ 8-10 blinks per minute, sleepy $=$ 4-6 blinks per minute), drowsiness can be identified \cite{Zeeshan}. Islam et al.~\cite{Arafat} calculate the eye aspect ratio to determine the eye closure time and total blink per minute. These values can then be compared to appropriate thresholds and an alarm is activated if the value exceeds or falls below the corresponding thresholds (depending on the critical variable to consider).

%A more advanced method is used to determine the drowsiness state of a driver \cite{Antoine}. 
%In \cite{Antoine} a more advanced method is developed to determine the drowsiness state of a driver. 
Picot et al.~\cite{Antoine} developed a more advanced method to determine the drowsiness state of a driver.
Their idea is analogous to the use of electro-oculograms (EOG) \cite{Galley}, where electrodes are placed near the eyes and the voltage signals are measured. They record visual signs from 60 hours of driving from different drivers. Based on data-mining techniques their algorithm then identifies patterns of drowsiness. Moreover, another similar approach uses a head gear to record the pupils on a driving simulation \cite{Takchito}. The algorithm computes the vertical length of dark pupils and is able to detect drowsiness from this variable.

A development in drowsiness detection uses binarization in combination with image filters. The system proposed by Ueno et al.~\cite{Hiroshi} is able to detect the vertical position of the eyes. Therein the algorithm takes into consideration the size of the eyes to calculate the ratio of opened and closed eyes. 

The drowsiness of the driver is addressed by detecting the state of the eyes when they are closed for a certain period of time \cite{Ioana,Pratiksha,Amin}. Therein, the relevant parts of the eyes are detected using Haar-Cascade classifiers \cite{Ioana,Pratiksha}. This approach seems particularly suitable to be easily integrated into a driver-assistance system, since it is merely based on eye detection. However, it only detects whether the driver has already fallen into the microsleep state, which may be too late for the successful prevention of road accidents. 

%From \cite{Zeeshan}, we are able to grasp the idea on comparing the number of blinking which is important to compare the speed rate of every blinking with a threshold. Another important variable is the eye aspect ratio which can be referred in \cite{Arafat}. As in \cite{Antoine,Takchito}, we need to simulate a scenario that imitates a sleepy driver, so that we can measure the blinking’s speed and compare it to a normal blinking. Furthermore, to determine the size of the eyes as in \cite{Hiroshi} is very crucial in this project because every person has their own significant eye’s size, thus it will affect the speed. To test the reliability of the project, we need to evaluate different scenario such as different head position that can effect the detection of the eye \cite{Amin}.

The goal of our work is to set up a comparably simple real-time drowsiness detection system with minimal requirements and to challenge it in an extensive evaluation by executing various test cases. 
In our work we follow Haq et al.~\cite{Zeeshan} and use the eye blinking rate to decide if a driver becomes drowsy. However, we determine the eye blinking rate differently. We measure the eye aspect ratio (used by Islam et al.~ \cite{Arafat}) and derive the eye blinking speed from it. In order to set the threshold for the eye blinking rate below which the driver is considered to be drowsy, we follow Picot et al.~\cite{Antoine} and Hayami et al.~\cite{Takchito} by simulating a scenario that imitates sleepiness of a driver. The threshold will also be dependent on the individual size of the eyes \cite{Hiroshi}. After fixing the threshold experimentally, we challenge our drowsiness detection method in a series of test cases (partly inspired by Suhaiman et al.~\cite{Amin}). While the proposed method works in principle, our extensive evaluation reveals a reduction of the reliability in certain scenarios (e.g.~tilting the head by larger angles). In this way we can systematically elaborate the next research steps to increase the accuracy of our simple drowsiness detection system in a broader range of scenarios.   

%From \cite{Zeeshan}, we are able to grasp the idea on comparing the number of blinking which is important to compare the speed rate of every blinking with a threshold. Another important variable is the eye aspect ratio which can be referred in \cite{Arafat}. As in \cite{Antoine,Takchito}, we need to simulate a scenario that imitates a sleepy driver, so that we can measure the blinking’s speed and compare it to a normal blinking. Furthermore, to determine the size of the eyes as in \cite{Hiroshi} is very crucial in this project because every person has their own significant eye’s size, thus it will affect the speed. To test the reliability of the project, we need to evaluate different scenario such as different head position that can effect the detection of the eye \cite{Amin}.

\section{Methods}
The drowsiness of a driver can be anticipated by analyzing the movement of the eyelids. The eyelids move slower than a normal blink. In this paper, we implement an algorithm that allows us to determine the speed of the blinking eye. Moreover, we use Histogram of Oriented Gradients (HOG) and a Linear Support Vector Machine (SVM) method to improve eye detection (4.89\% higher accuracy compared to Haar-Cascade, see Rahmad et al.~\cite{Rahmad}).

%In this section, two methods will be explained in detail. The first method is eyes detection and the second is the calculation of the eye blinking speed.

\subsection{Eyes Detection}
The algorithm is trained to detect the landmarks of the facial features in the dlib library by using an ensemble of regression tress \cite{Christos}. HOG image descriptor and SVM are the method for the process training of an object \cite{Navneet}. There are many datasets available to detect these landmarks and we are using the dataset from IBUG which has 68 points of facial landmarks \cite{Vahid}.

\subsection{Eye Blinking Speed}

\begin{figure}[t]
  \centering
  \subfloat[6 points eye landmarks ]{\includegraphics[width=0.492\textwidth]{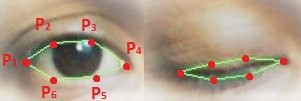}\label{6 points}}
  \hfill
  \subfloat[EAR over time for an eye blinking around frame 110]{\includegraphics[width=0.49\textwidth]{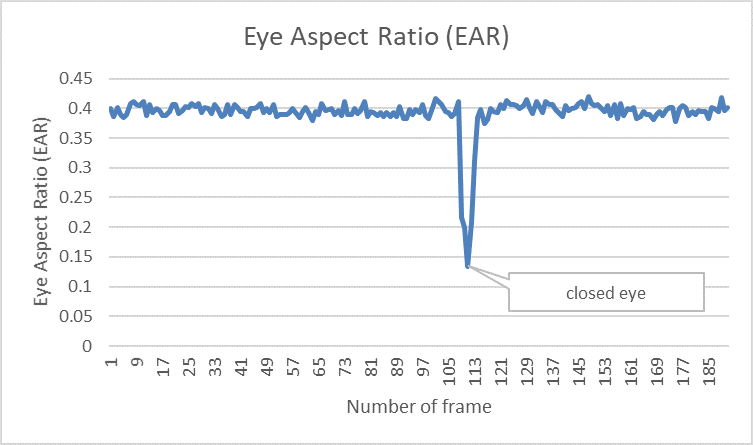}\label{EAR}}
  \caption{Comparison between opened and closed eye \cite{Tereza}}
\end{figure}

Having detected the eyes of the driver, the next step is to determine the eye blinking speed. 
Firstly, we have to detect whether the state of the eye is opened or closed.
%Therefore, it is compulsory to calculate the eye aspect ratio (EAR), following \cite{Tereza}, by}
A suitable measure to derive the state of the eye is the eye aspect ratio (EAR). We follow the definition of Soukupov\'a et al.~\cite{Tereza}: 
\begin{equation} \label{Eq: Definition EAR}
    \text{EAR} = \frac{|p_2-p_6|+|p_3-p_5|}{2|p_1-p_4|} \ ,
\end{equation}
where $p_1$ to $p_6$ are the facial landmarks as depicted in \autoref{6 points}. %In \autoref{Sec: Evaluation}, we determine typical values for the EAR for opened and closed eyes. 
When the eye is opened, the EAR is above 0.35, but when it is closed, the value rapidly dropped below 0.15 (see \autoref{EAR} and \autoref{Sec: Evaluation} for the corresponding experiments).

\begin{figure}[t]
\includegraphics[width = 0.6\textwidth]{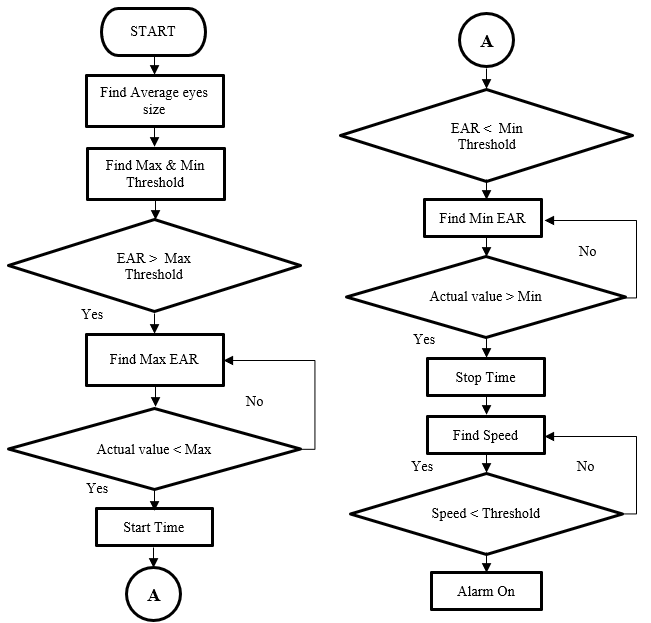}
\centering
\caption{Flow diagram to determine the eye blinking speed}
\centering
\label{flow}
\end{figure}

Based on the flow diagram in \autoref{flow}, firstly, the algorithm will find the average eye size (AES) of the driver defined by
\begin{equation} \label{Eq: AES}
    \text{Average Eye Size (AES)} = \frac{\text{Max EAR1} + \text{Max EAR2} + \text{Max EAR3}}{3} \ ,
\end{equation}
i.e.~we take the arithmetic mean of three measured maximum EARs. (Notice that the accuracy of the AES can be improved by measuring more maximum EARs, but this is at the expense of a higher computational effort.)

Afterwards, Max and Min Threshold will be calculated based on the average eye's size value. Max and Min Threshold are defined by
\begin{align} \label{Eq: Max Thresh}
    \text{Max Threshold} &= {\frac{2}{3}}\text{AES} + 0.0467  \ , \\
 \label{Eq: Min Thresh}
    \text{Min Threshold} &= \text{Max Threshold} - {0.05} \ .
\end{align}
The numerical values in these equations were found empirically. 
After Max and Min Threshold have been determined, the algorithm will search for the maximum value of the EAR (denoted by $\text{Max EAR}$) while capturing the images frame by frame. When the current EAR is less than the current maximum value, it will start the timer and at the same time find the minimum value of the EAR (denoted by $\text{Min EAR}$). The final minimum value is determined, when the current EAR eventually becomes larger than the minimum value, thus the timer will stop. The blinking speed for each blink can be calculated by 
\begin{equation} \label{Eq: Blinking speed}
    \text{Blinking Speed} = \frac{\text{Max EAR} - \text{Min EAR}}{\text{Start Time} -\text{Stop Time}} \ .
\end{equation}
%The speed is calculated for each blink. 
If the blinking speed becomes sufficiently low, the algorithm will activate an alarm system. For this purpose we introduce an empirically determined drowsiness threshold. Whenever the eye blinking speed is below this drowsiness threshold, the algorithm identifies the driver as being in a drowsy state. 
%If the speed drops below some experimentally determined the Sleepy Threshold, this indicates that the driver is sleepy.  Hence, the algorithm will activate an alarm system.

\section{Evaluation} \label{Sec: Evaluation}

For the evaluation of our drowsiness detection system we proceed as follows. We first show experimentally that changing the distance between eyes and camera leaves the EAR invariant. Then, we determine the speed of the eye blinking in the wakeful and sleepy state, respectively. Finally, we evaluate the impact of changes in the head positions on our system.  
%This section has four parts. The first part is about the relationship between EAR and the distance between the eye and camera. The second and third part evaluate the speed of the blinking for normal and slow blink. Lastly, the effects of different head positions on the detection of the eye.

\subsection{Evaluation of the Impact of Distance Variations on the EAR} \label{Evaluation for constant EAR}

\begin{figure}[t]
  \centering
  \subfloat[The distance between $p_2$ and $p_6$]{\includegraphics[width=0.492\textwidth]{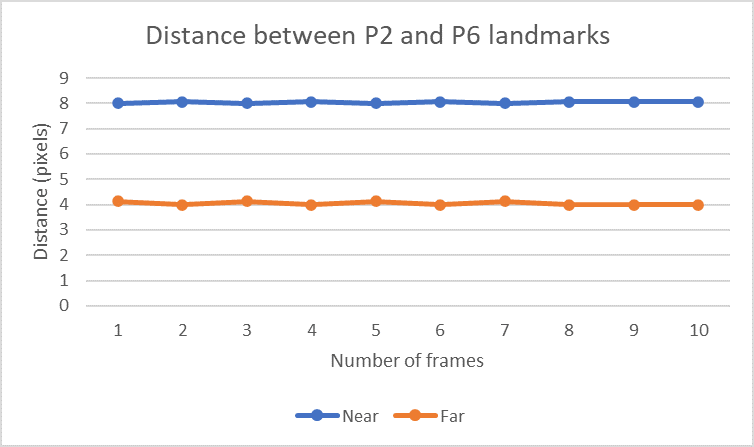}\label{EAR3}}
  \hfill
  \subfloat[EAR from both images]{\includegraphics[width=0.49\textwidth]{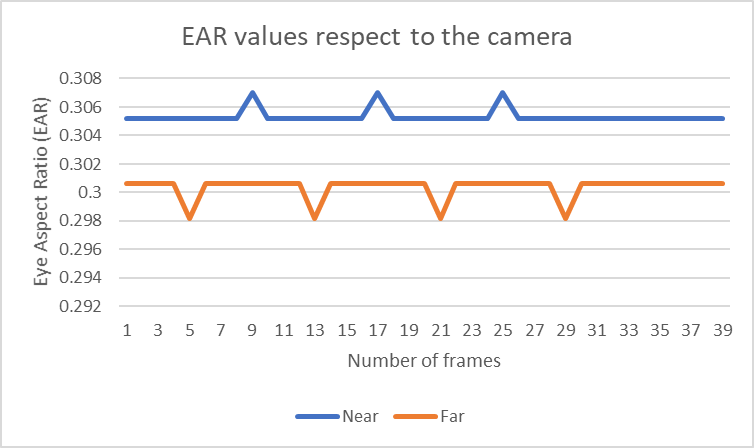}\label{EAR4}}
  \caption{Result for constant EAR's evaluation}
\end{figure}

%It is important to have a constant EAR in this project. This is due to the size of the eyes changes when there is an alteration of distance between the eyes and the camera. 
When alternating the distance between eyes and camera, the apparent size of the eyes will change. However, based on \cref{Eq: Definition EAR} we expect the EAR to remain invariant. 
%For instance, the EAR should remain the same whether the distance of the eyes is 30 cm or 60 cm from the camera. 
%This can be proven by running a test on the algorithm.

The following evaluation shows experimentally that the EAR is indeed invariant under modification of the distance of camera and eyes. For this purpose, two similar images with different sizes are used instead of a live stream video. The reason is to have a fixed EAR reference from a static image, enabling a comparison of the EAR from both images that have different eye's sizes.
The sizes of the eyes in both images are relatively different because one image is close to the camera and the other is far from it. The size can be measured by calculating the difference distance between two points of facial landmarks (e.g.~$p_2$ and $p_6$ in \autoref{EAR3}, which are the upper and lower eyelid) in both images. 

The distance between these points in Image 1 (Near, blue line in \autoref{EAR3} and \autoref{EAR4}) is two times bigger than Image 2 (Far, orange line in \autoref{EAR3} and \autoref{EAR4}) which is approximately 8 and 4 pixels (see \autoref{EAR3}), respectively. It shows that the position of the eyes in Image 1 is nearer to the camera than Image 2.

The ratio between both distances is approximately 1:2. The result in \autoref{EAR4} shows that the measured EARs for Images 1 and 2 are approximately 0.3052 and 0.3006, respectively. The deviation is only $1.5\%$, which is acceptable for our purposes.

\subsection{Evaluation for Normal Blinking}
The purpose of this evaluation is to check whether Max EAR, Min EAR, and hence, the blinking speed in \cref{Eq: Blinking speed} are calculated correctly. Moreover, the average blinking speed in the wakeful state can be determined from this test.

The participants were asked to blink normally for 8 times. The first three blinks were analyzed by the algorithm to obtain the average size of the eyes. The other five blinks were necessary to evaluate the EAR and the speed of the blinking.
Three tests from three participants were conducted thoroughly and the results are as follows.

%edit the subsubsection position
%change the images on the top
\subsubsection*{Participant  1:}

\begin{figure}[t]
  \centering
  \subfloat[Speed of normal blinking]{\includegraphics[width=0.5\textwidth]{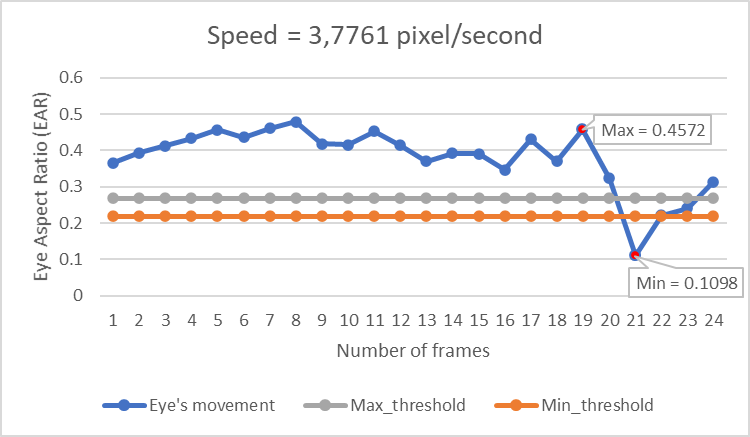}\label{Normal1}}
  \hfill
  \subfloat[Speed values from five blinks]{\includegraphics[width=0.485\textwidth]{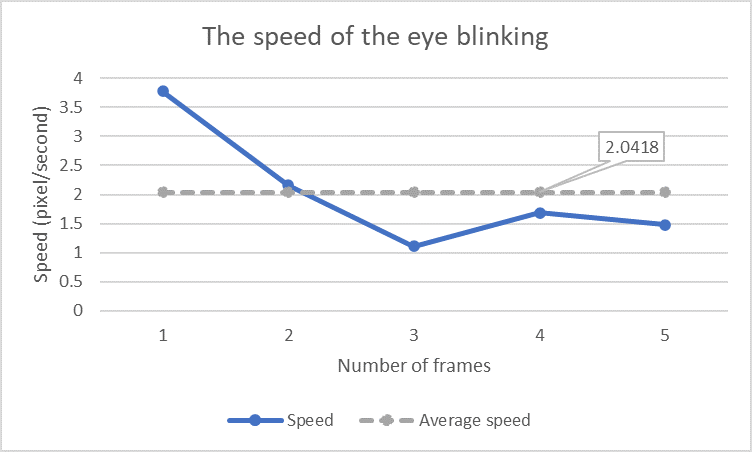}\label{Normal2}}
  \caption{Result from Participant 1 (normal blink)}
\end{figure}
%The first participant was instructed to blink normally eight times while looking at the camera.

\autoref{Normal1} shows that the maximum of the EAR, which is the highest value of the EAR before the eyes start to close, is calculated correctly above the Max Threshold (defined in \eqref{Eq: Max Thresh}). Moreover, we see that there is only one data point below the minimum threshold (orange line). The minimum threshold can be determined using \eqref{Eq: Min Thresh}.
We experimentally obtain $\text{Max~EAR}=0.4572$ and $\text{Min~EAR}=0.1098$.
\autoref{Normal2} shows that the values of the blinking speed are above 1 pixel/second and the average speed is 2.0418 pixel/second.

\subsubsection*{Participant  2:}
\begin{figure}[t]
  \centering
  \subfloat[Normal blinking’s speed]{\includegraphics[width=0.49\textwidth]{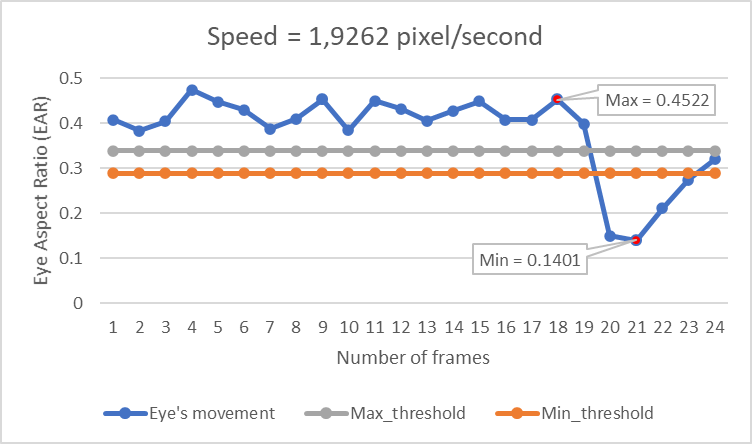}\label{Normal3}}
  \hfill
  \subfloat[Speed values from five blinks]{\includegraphics[width=0.492\textwidth]{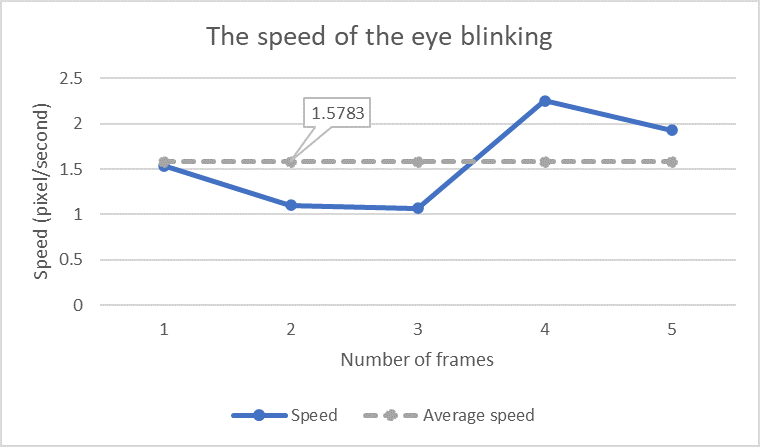}\label{Normal4}}
  \caption{Result from Participant 2 (normal blink)}
\end{figure}

%The second participant was instructed to blink normally eight times while looking at the camera\newline

\autoref{Normal3} shows that $\text{Max~EAR}=0.4522$. However, there are two data points below the minimum threshold, whose difference in the EAR is 0.01 pixels. The algorithm needs to determine which value to choose as a minimum. The final minimum value is $\text{Min~EAR}=0.1401$ which is the correct value because this is the value when the eyes are completely closed. If the differences between the values below minimum threshold is less than 0.01, then the final minimum will be the first value because this is the point where the eye is completely shut.\newline
We can see that \autoref{Normal4} is similar to \autoref{Normal2} where the speed values are above 1 pixel/second and the average speed is 1.5783 pixel/second.

\subsubsection*{Participant  3:}
\begin{figure}[t]
  \centering
  \subfloat[Normal blinking’s speed]{\includegraphics[width=0.49\textwidth]{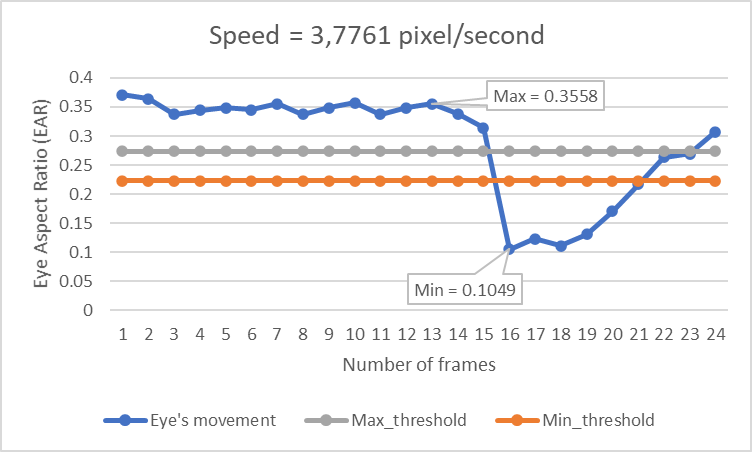}\label{Normal5}}
  \hfill
  \subfloat[Speed values from five blinks]{\includegraphics[width=0.492\textwidth]{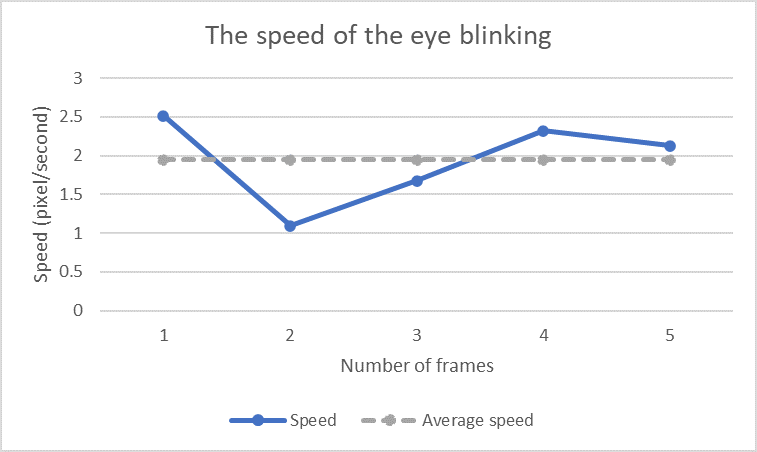}\label{Normal6}}
  \caption{Result from Participant 3 (normal blink)}
\end{figure}

%The third participant was instructed to blink normally eight times while looking at the camera\newline

In this case the EAR reaches a rather flat maximum at $\text{Max~EAR}=0.3558$ before dropping sharply below the minimum threshold (see \autoref{Normal5}). The minimum EAR is measured to be $\text{Min~EAR}=0.1049$. Again, the blinking speed never drops below 1 pixel/second (see \autoref{Normal6}). The average eye blinking rate is approximately at 2 pixels/second. 
\bigskip 

To summarize the results of the three tests, the average speed for normal blinking has a threshold value of 1 pixel/second. However, there are certain cases where the eye blinking speed can slightly drop below this threshold.

\subsection{Evaluation for Sleepy Blinking}

For the measurement of the EARs and the eye blinking rate in the drowsy state, we essentially repeat the previous experiments.
%This part is very crucial because it tests the main objective of this project whether it is capable to solve the problem stated in the introduction. 
The evaluation was conducted by testing three participants with 5 trials. They were asked to imitate the behavior of a sleepy driver in front of the camera by closing their eyes slowly. The corresponding data are shown and discussed below.

\subsubsection*{Participant  1:}
\begin{figure}[t]
  \centering
  \subfloat[Sleepy blinking’s speed]{\includegraphics[width=0.49\textwidth]{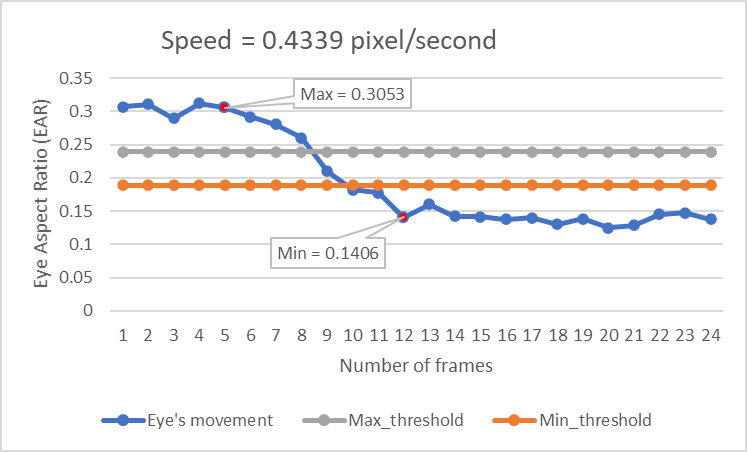}\label{Slow1}}
  \hfill
  \subfloat[Speed values from five blinks]{\includegraphics[width=0.492\textwidth]{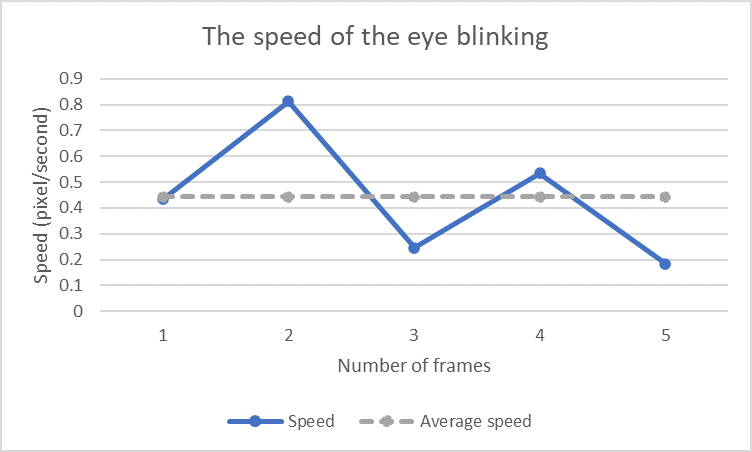}\label{Slow2}}
  \caption{Result from Participant 1 (sleepy blink)}
\end{figure}

%{The fourth participant was instructed to blink normally three times and blink slowly  five times to imitate a sleepy driver while looking at the camera\newline

In \autoref{Slow1} it can be seen that the number of frames from $\text{Max~EAR}=0.3053$ down to $\text{Min~EAR}=0.1406$ is larger than in the wakeful state, indicating a lower eye blinking rate. 
%In \autoref{Slow1}, it can be seen that the line graph from max (0.3053) until min (0.1406) has more number of frames compare to the normal blinking in chapter 4.2. This demonstrates the time taken from max to min is longer.
Indeed, \autoref{Slow2} shows the smaller speed values for all five blinks. The average speed is 0.4421 pixel/second.

\subsubsection*{Participant  2:}
\begin{figure}[t]
  \centering
  \subfloat[Sleepy blinking’s speed]{\includegraphics[width=0.49\textwidth]{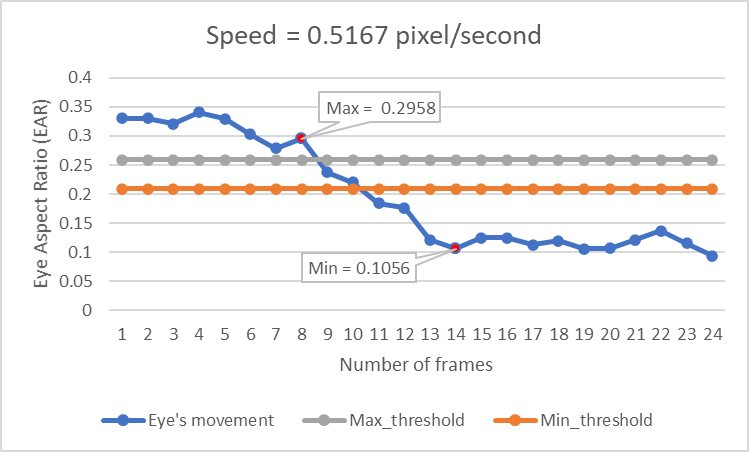}\label{Slow3}}
  \hfill
  \subfloat[Speed values from five blinks]{\includegraphics[width=0.492\textwidth]{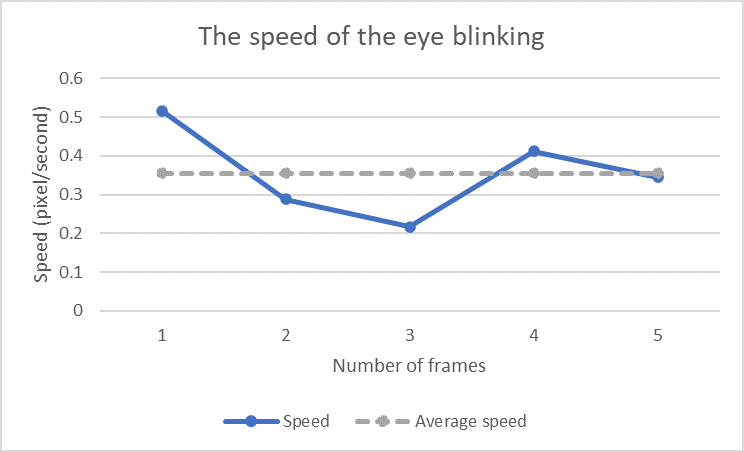}\label{Slow4}}
  \caption{Result from Participant 2 (sleepy blink)}
\end{figure}

%{The fifth participant was instructed to blink normally three times and blink slowly  five times to imitate a sleepy driver while looking at the camera\newline

\autoref{Slow3} has a slightly different result from the first test (see \autoref{Slow1}), specifically regarding the values below the minimum threshold. It can be seen that the EAR remains below the minimum threshold after $\text{Min~EAR}=0.1056$ has been reached. 
%there are many values after the final min. 
The participant was closing his eyes longer than usual which mimics one of the main behaviors of a sleepy person. The values are in the range between 0.5167 pixel/second and 0.2172 pixel/second. The average speed value is 0.3557 pixel/second.

\subsubsection*{Participant  3:}
\begin{figure}[t]
  \centering
  \subfloat[Sleepy blinking’s speed]{\includegraphics[width=0.49\textwidth]{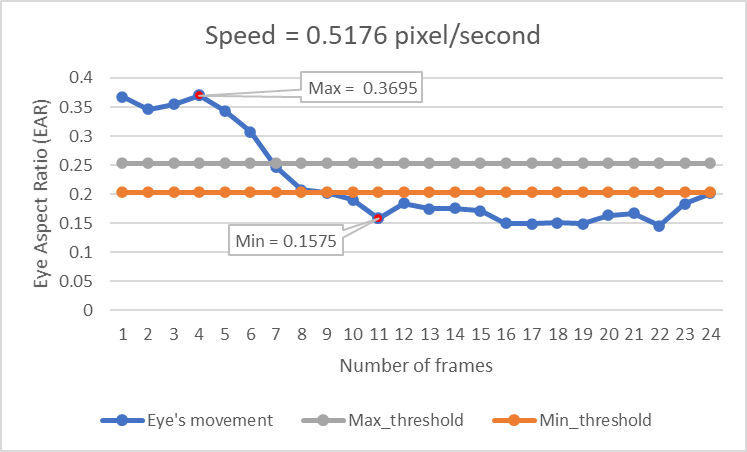}\label{Slow5}}
  \hfill
  \subfloat[Speed values from five blinks]{\includegraphics[width=0.492\textwidth]{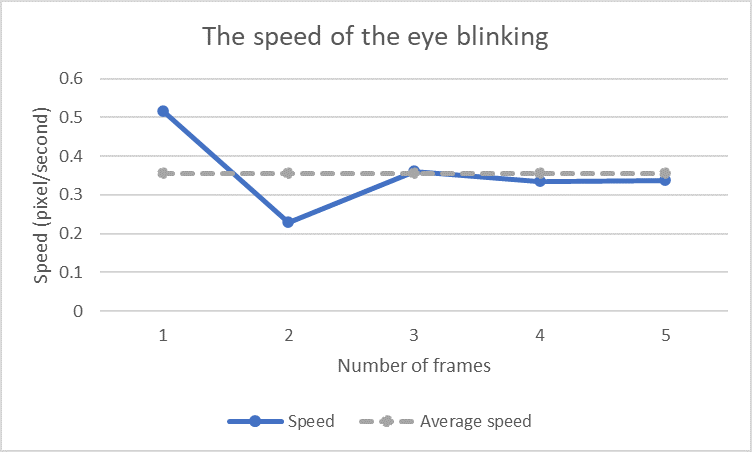}\label{Slow6}}
  \caption{Result from Participant 3 (sleepy blink)}
\end{figure}

%{The sixth participant was instructed to blink normally three times and blink slowly  five times to imitate a sleepy driver while looking at the camera\newline

The graph in \autoref{Slow5} has a similar pattern as in the second test (\autoref{Slow3}), where there are many values below the minimum threshold.
The results from the third test (\autoref{Slow6}) came out as expected and the speed values are in the range between 0.5176 pixel/second and 0.2292 pixel/second. The average speed of the eye blinking is 0.35586 pixel/second.
\bigskip 

As a result of these tests, it can be concluded that the average speed value for sleepy eye blink is below 0.5 pixel/second and hence the drowsiness threshold to activate the alarm will be set to 0.55 pixel/second (taking into account a safety buffer). Thus, this is the threshold value which will activate the alarm if the speed drops below it.

\subsection{Evaluation for Different Head Positions}

In this last evaluation, the participant went through different situations to test the reliability of the system. These test cases, inspired by Suhaiman et al.~\cite{Amin}, simulate the scenario when the driver moves his head in different directions.

In the first test case, the participant was instructed to move his head upwards, downwards, left and right while looking in front. Eventually, in the second test case, the participant also move his head in the same direction as in the first situation but the eyes also follow the direction of the head. For example, if the participant tilts his head upwards, his eyes should look upwards. \autoref{Table:Evaluation of different head positions} shows the different situations of the head movement and the results.

\begin{table}
\caption{Evaluation of different head positions}
\label{Table:Evaluation of different head positions}
\begin{tabular}{ |p{3cm}|p{10cm}|}
\hline
\multicolumn{2}{|c|}{\textbf{Eyes look in front}} \\
\hline
\textbf{Test case} & \textbf{Result} \\
\hline
Head faces upward and downward & 
• Able to detect eyes up to a certain degree\newline
• Able to detect eye blinking\newline
• EAR becomes smaller\newline
• Smaller EAR affects finding the correct Max EAR and Min EAR\\
\hline
Head turns to the left and right &  
• Able to detect eyes as long both eyes are visible to the camera \newline
• Able to detect eye blinking but the speed is inaccurate \newline
• Some speeds cannot be calculated for certain head poses\\
\hline
\multicolumn{2}{|c|}{\textbf{Eyes follow head’s movement}} \\
\hline
\textbf{Test case} & \textbf{Result} \\
\hline
Head faces upward and downward & 
• Able to detect eyes up to a certain degree \newline
• EAR becomes smaller (bigger than when eyes look in front) \newline
• Able to detect eye blinking \newline
• Speed is less accurate\\
\hline
Head turns to the left and right &  
• Able to detect eyes as long both eyes are visible to the camera \newline
• Able to detect eye blinking but the speed is inaccurate \newline
• Some speeds cannot be calculated for certain head poses\\
\hline
\end{tabular}
\end{table}

\section{Conclusion}
In this paper, we have shown that by calculating the speed of the eye blinking, we are able to distinguish between a wakeful and a drowsy blink of a driver in real-time. In particular, we can also detect the gradual process of a driver becoming drowsy. Such a real-time drowsiness detection system plays a key role in preventing car accidents due to microsleep. 

However, our extensive evaluations also revealed some deficits, which should be addressed in future developments of our simple algorithm. 
Firstly, the problem where the algorithm cannot detect the eyes in a certain angle when the head is in a certain position (such as tilted upwards or downwards) can be improved either by identifying the rotation of the head and give conditions in the program or by including additional cameras positioned in different angles \cite{cabin}. 
Secondly, facial expression such as smiling has an impact on the measured EAR. Therefore an additional algorithm is needed to detect facial expression, which can then be used to adapt the maximum and minimum thresholds defined in \eqref{Eq: Max Thresh} and \eqref{Eq: Min Thresh}. Another improvement of our algorithm regards the inclusion of optical effects that can occur due to eye glasses.

\bibliographystyle{splncs}
\bibliography{egbib}
\end{document}